\documentclass[runningheads]{llncs}
\usepackage{amsmath, amssymb}
\usepackage{graphicx}    
\usepackage{wrapfig}
\usepackage{multirow}
\usepackage{makecell}
\usepackage{colortbl}
\usepackage{xcolor}
\usepackage{tikz}
\usepackage{makecell}
\usepackage{multirow}
\usepackage{xcolor}
\usepackage{amsmath}     
\usepackage{booktabs}    
\usepackage{pifont}  
\usepackage[table]{xcolor}
\usepackage{colortbl}    

\makeatletter
\def\blfootnote{\gdef\@thefnmark{}\@footnotetext}
\makeatother
\definecolor{oursrow}{RGB}{242,247,242} 

\usepackage[final,year=2026,ID=***]{sty}

\usepackage{sty}

\usepackage[accsupp]{axessibility}  

\definecolor{LightCyan}{rgb}{0.88,1,1}
\definecolor{cvprblue}{rgb}{0.21,0.49,0.74}
\usepackage[pagebackref,breaklinks,colorlinks,allcolors=cvprblue]{hyperref}

\usepackage{graphicx}
\usepackage{booktabs}
\usepackage{fix-cm}
\usepackage{bm}
\usepackage{multirow}
\usepackage{longtable}
\usepackage{supertabular}
\usepackage{colortbl}
\usepackage{subcaption}
\usepackage[most]{tcolorbox}
\usepackage[dvipsnames]{xcolor}

\usepackage{orcidlink}

\begin{document}

\title{LIVE: Leveraging Image Manipulation Priors for Instruction-based Video Editing}
\titlerunning{LIVE}

\authorrunning{Wang et al.}
\author{
\textbf{Weicheng Wang$^{1*}$, Zhicheng Zhang$^{1*}$, Zhongqi Zhang$^{1}$, Juncheng Zhou$^{1}$,} 
\textbf{Yongjie Zhu$^{3\dag}$, Wenyu Qin$^3$, Meng Wang$^3$, Pengfei Wan$^3$, Jufeng Yang$^{124\ddag}$}\\
} 

\institute{{\small $^1$ Nankai University}
{\small $^2$ Pengcheng Laboratory}
{\small $^3$ Kuaishou Technology} \\
{\small $^4$ Nankai International Advanced Research Institute (SHENZHEN·FUTIAN)}\\
{\small
\texttt{wangweicheng777@mail.nankai.edu.cn}, \texttt{gloryzzc6@sina.com}}\\
{\small
\texttt{zzhongqi37@gmail.com}, \texttt{zhou\_juncheng@mail.nankai.edu.cn}}\\
{\small
\texttt{\{zhuyongjie,qinwenyu,wangmeng46,wanpengfei\}@kuaishou.com}
} \\
{\small
\texttt{yangjufeng@nankai.edu.cn}
}}

\blfootnote{$*$: Equal Contribution. $\dag$: Project Leader.  $\ddag$: Corresponding Author.}

{
\maketitle
\begin{center}
    \centering
    \captionsetup{type=figure}
    \includegraphics[width=1.0\textwidth]{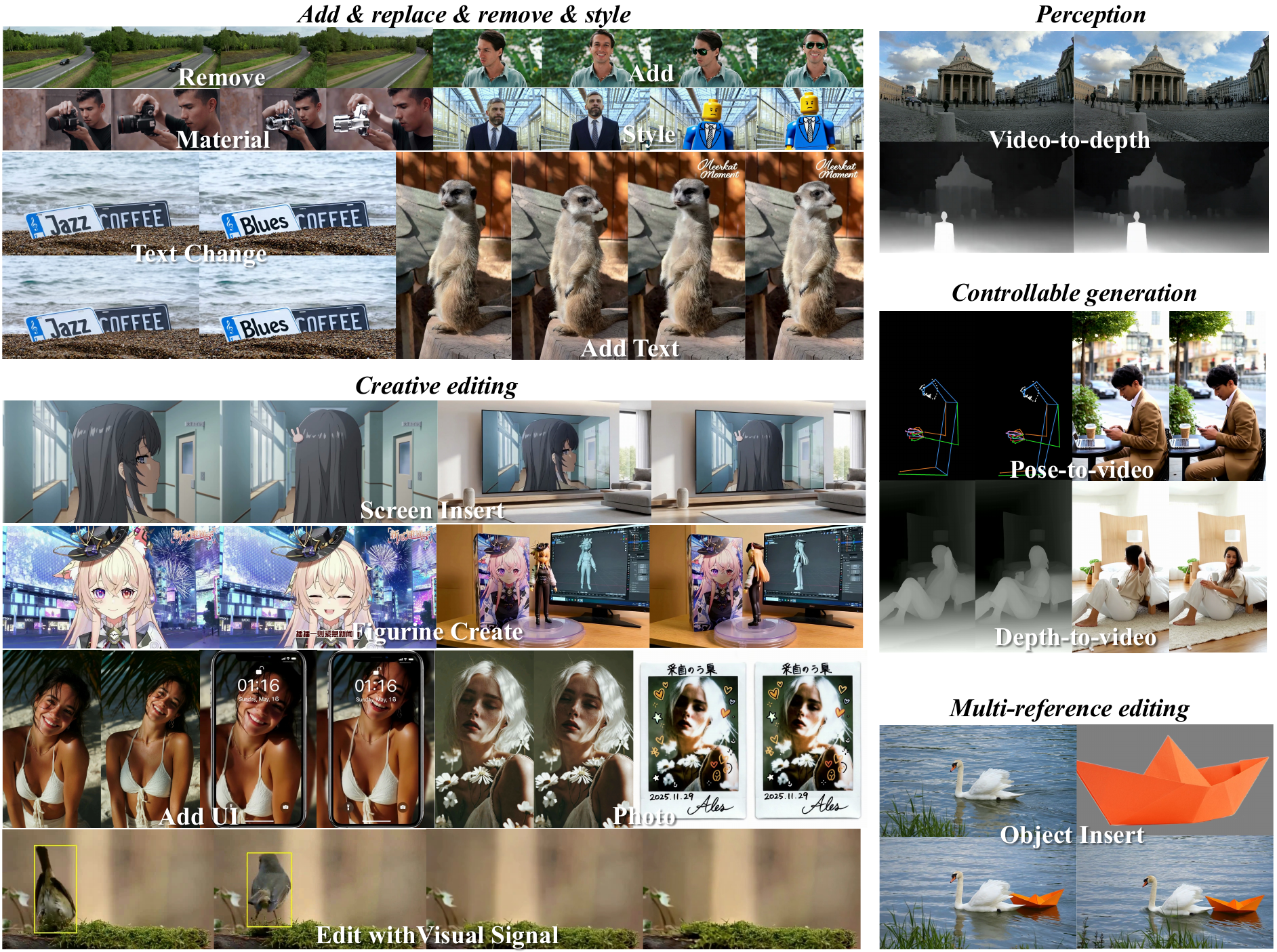}
    \vspace{-0.4cm}
    \captionof{figure}{Performance of LIVE in various video editing tasks. In addition to common video editing capabilities, our method enables creative video editing tasks. Furthermore, we support video input at different resolution ratios.}
    \label{fig:teaser}
\end{center}
}

\begin{abstract}
Video editing aims to modify input videos according to user intent.
Recently, end-to-end training methods have garnered widespread attention, constructing paired video editing data through video generation or editing models. 
However, compared to image editing, the high annotation costs of video data severely constrain the scale, quality, and task diversity of video editing datasets when relying on video generative models or manual annotation.
To bridge this gap, we propose LIVE, a joint training framework that leverages large-scale, high-quality image editing data alongside video datasets to bolster editing capabilities.
To mitigate the domain discrepancy between static images and dynamic videos, we introduce a frame-wise token noise strategy, which treats the latents of specific frames as reasoning tokens, leveraging large pretrained video generative models to create plausible temporal transformations.
Moreover, through cleaning public datasets and constructing an automated data pipeline, we adopt a two-stage training strategy to anneal video editing capabilities. 
Furthermore, we curate a comprehensive evaluation benchmark encompassing over 60 challenging tasks that are prevalent in image editing but scarce in existing video datasets. 
Extensive comparative and ablation experiments demonstrate that our method achieves state-of-the-art performance.
\textbf{The source code will be publicly available.}
\end{abstract}
    
\section{Introduction}
Instruction-based video editing~\cite{wu2025insvie,chen2025ivebench} aims to synthesize a target video by transforming input videos according to specific user instructions. 
This facilitates various real-world applications, such as personalized video creation~\cite{xing2024make} and digital human animation~\cite{xu2026anchorcrafter}. 
Despite the recent breakthrough in text-to-video (T2V)~\cite{wang2023modelscope,yang2024cogvideox} generation and image editing models~\cite{brooks2023instructpix2pix,liu2025step1x,sheynin2024emu}, extending these advances to video editing remains a non-trivial challenge, executing precise semantic modifications while ensuring high-fidelity preservation in both spatial and temporal dimensions.

\begin{figure*}[!tbp]
  \centering
   \includegraphics[width=1.0\textwidth]{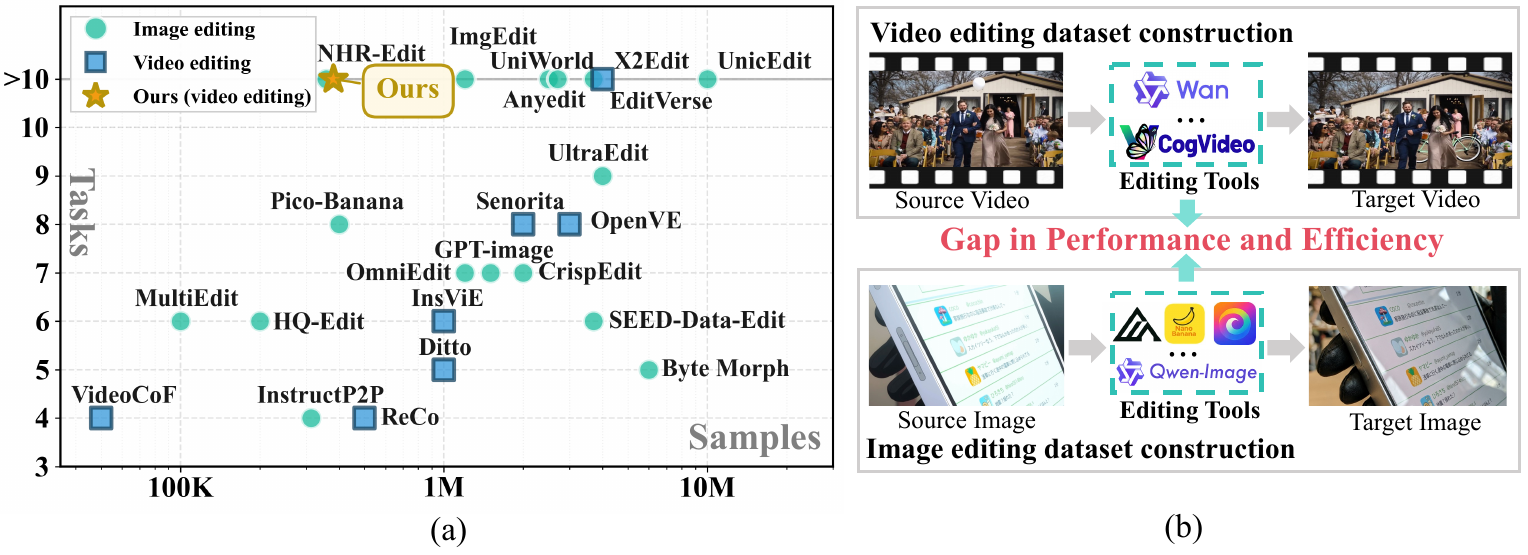} 
   \vspace{-10pt}
  \caption{(a) Comparison of task coverage and training data scale among representative instruction-based image and video editing methods. Our method achieves more editing tasks with less video training data. (b) with advanced editing tools, the construction of image editing datasets achieves higher feasibility and quality compared to video counterparts.}
  \label{fig:motivation}
\end{figure*}

Existing paradigms for video editing can be broadly categorized into training-free and end-to-end training-based approaches. 
Training-free methods~\cite{ceylan2023pix2video,wang2023zero,qi2023fatezero} rely on zero-shot techniques to perform editing with pre-trained generative models. 
Representative approaches include TokenFlow~\cite{geyer2023tokenflow} and AnyV2V~\cite{ku2024anyv2v}, which leverage DDIM inversion~\cite{song2020denoising} to reconstruct target latents.
Instead, FlowDirector~\cite{li2025flowdirector} adopts inversion-free strategies based on flow matching.
However, without training data, these methods suffer from longer inference times and limit to easy editing scenarios.
To overcome these limitations, end-to-end training frameworks have recently garnered increasing attention. 
For training, a key challenge is the scarcity of high-quality, paired video editing data.
Consequently, many efforts focus on automated data synthesis pipelines to generate large-scale video pairs. 
Early, InsV2V~\cite{cheng2023consistent} employs the Prompt2Prompt~\cite{hertz2022prompt} to create data by altering text prompts. 
InsViE-1M~\cite{wu2025insvie} generates target videos by editing the first frame and propagating changes through DDIM inversion across the temporal dimension.
Recently, OpenVE-3M~\cite{he2025openve} and Ditto~\cite{bai2025scaling} introduce control signals to guide controllable video models, producing editing pairs with higher quality.
Despite persistent efforts to enhance data quality, instruction-based video editing remains constrained by the scarcity of large-scale, diverse datasets compared to the image editing landscape.
This data gap makes video editing struggle with complex manipulations like Nano Banana Pro (NBP) or Seedream 4.0.
The bottleneck is two-fold. 
First, current video editing datasets suffer from a lack of task diversity. 
As illustrated in Fig.~\ref{fig:motivation} (a), most existing methods support fewer than 10 tasks, whereas recent image editing frameworks such as UnicEdit~\cite{ye2025unicedit} leverage over 10M pairs to accommodate 22 distinct tasks. 
Second, scaling video editing tasks introduces costs and quality control challenges.
On one hand, manually constructing video editing data via human annotation not only requires professional editing expertise but also incurs substantial time costs. 
Alternatively, a feasible approach is to leverage existing video generative models to build an automated synthesis pipeline.
However, as shown in Fig.~\ref{fig:motivation} (b), relying on open-source generators like Wan-VACE~\cite{jiang2025vace} often yields samples with lower quality and higher filtering complexity compared to proprietary counterparts like NBP, while requiring more inference overhead for temporal modeling.

To overcome these limitations, a promising direction is to extend image editing capabilities into the video domain.
Specifically, we propose \textbf{LIVE} to \textbf{L}everaging \textbf{I}mage editing priors for
Instruction-based \textbf{V}ideo \textbf{E}diting.
First, we establish cross-domain data alignment. 
Given that images lack the temporal information, we expand image pairs into pseudo-videos by repeating them along the temporal dimension during training.
However, simply replicating may lead to a static bias, where it fails to learn dynamic transitions. 
To mitigate this, we introduce a Frame-wise Token Noise Strategy. 
Specifically, we inject noise into latent tokens of random frames within the input pseudo-videos.
By masking these tokens in optimization, we force the model to implicitly reason latent temporal dynamics at noise-corrupted positions.
Second, we adopt a two-stage training paradigm. 
In the initial stage, we utilize a high-ratio mixture of image-editing data to instill diverse transformation capabilities into the Diffusion Transformer (DiT). 
In the subsequent stage, we reduce the proportion of image data, shifting the focus towards refining temporal coherence.
As shown in Fig.~\ref{fig:teaser}, LIVE demonstrates powerful performance on multiple fine-grained editing tasks.
Finally, to rigorously evaluate models across diverse and challenging scenarios, we introduce a benchmark LIVE-Bench. 
LIVE-Bench encompasses over \textbf{60} sub-tasks, specifically including novel capabilities demonstrated by proprietary image models like NBP, as well as complex editing tasks absent from current open-source video datasets.

Our contributions are as follows: 1) We propose a hybrid training framework that integrates both image and video editing data, facilitating the effective transfer of static editing knowledge to the dynamic video domain.
2) A comprehensive benchmark comprising over 60 diverse tasks is established. Extensive experiments validate that our approach attains state-of-the-art performance across various video editing scenarios.
\section{Related Work}
\subsection{Instruction-based Video Editing Models}
Instruction-based video editing~\cite{he2025openve,xia2025dreamve,zhang2025instructvedit} aims to transform input videos into a target output according to text instructions.
With the rapid advancement of generative models~\cite{rombach2022high,podell2023sdxl,wan2025wan,hong2022cogvideo}, the performance of video editing has significantly improved.
Current methodologies broadly fall into two paradigms: training-free techniques and learning-based approaches.
Based on pre-trained text-to-image or video models, training-free methods~\cite{qi2023fatezero,geyer2023tokenflow,flatten,kara2024rave} typically reconstruct the target latent by coupling DDIM Inversion with attention-feature manipulation.
However, heavily relying on zero-shot inference, these approaches often struggle in complex scenarios, frequently suffering from temporal inconsistencies or failing to execute substantial structural modifications.
To overcome these bottlenecks, recent efforts prioritize end-to-end optimization using large-scale supervised data. 
For instance, InstV2V~\cite{cheng2023consistent}, InsViE-1M~\cite{wu2025insvie}, Ditto~\cite{bai2025scaling}, and OpenVE~\cite{he2025openve} utilize extensive synthesized paired data to train the video editor, significantly enhancing their robustness.
Beyond data scaling, recent advances also explore architectural improvement.
UniVideo~\cite{wei2025univideo} and Tele-Omni~\cite{liu2026tele} integrate MLLMs for instruction comprehension, while Editverse~\cite{ju2025editverse} proposes a unified autoregressive framework for training image and video editing.
In this work, we aim to train a DiT-based instruction-guided video editing model, jointly leveraging both image and video editing data.

\subsection{Instruction-based Editing Datasets}
The construction of instruction-based editing datasets typically involves generating edit pairs using advanced editing models, followed by quality filtering.
With the advancement of image editing tools and MLLMs, the diversity, scale, and quality of image editing datasets have been significantly improved.
For instance, GPT-Image-Edit-1.5M~\cite{wang2025gpt} leverages GPT-4o to refine existing image editing datasets, enhancing editing quality.
UnicEdit-10M~\cite{ye2025unicedit} employs FLUX.1-Kontext~\cite{labs2025flux} and Qwen-Image-Edit~\cite{wu2025qwen} to create 2M data with 22 tasks, and train an MLLM for filtering.
Pico-banana-400K~\cite{qian2025pico} utilizes Nano Banana to synthesize editing pairs characterized by greater complexity and creativity.
In contrast, existing pipelines for video data synthesis are generally less robust than image models, often exhibiting lower success rates in producing high-quality editing pairs.
For example, InsV2V~\cite{cheng2023consistent} and InsViE-1M ~\cite{wu2025insvie} employ training-free techniques to synthesize training data.
To enhance consistency and quality, ReCo~\cite{reco}, OpenVE-3M~\cite{he2025openve} and Ditto~\cite{bai2025scaling} incorporate control signals into Wan-VACE to generate editing pairs.
However, the task diversity of them remains limited. 
Although Editverse~\cite{ju2025editverse} introduces more than 20 video editing tasks, it lacks complex tasks like NBP.
Furthermore, for video quality verification, MLLMs still lag behind images in terms of performance and cost. 
Therefore, we aim to leverage massive amounts of image editing data to transfer diverse editing capabilities to the video domain.

\section{Image-Video Dataset Preparation}

\begin{figure*}[t]
    \centering
    
    \definecolor{ImgHeaderBG}{RGB}{211, 234, 233}  
    \definecolor{VidHeaderBG}{RGB}{224, 243, 218}  
    \definecolor{ImgBase}{RGB}{211, 234, 233}
    \definecolor{VidBase}{RGB}{224, 243, 218}

    \newcommand{\ditem}[3]{%
        \tikz[baseline=-0.2em] \fill [#1] (0,0) rectangle (0.6em,0.6em); \textbf{#2} \textcolor{gray}{(#3)}%
    }

    \begin{minipage}{0.35\textwidth}
        \centering
        \includegraphics[width=1.0\textwidth]{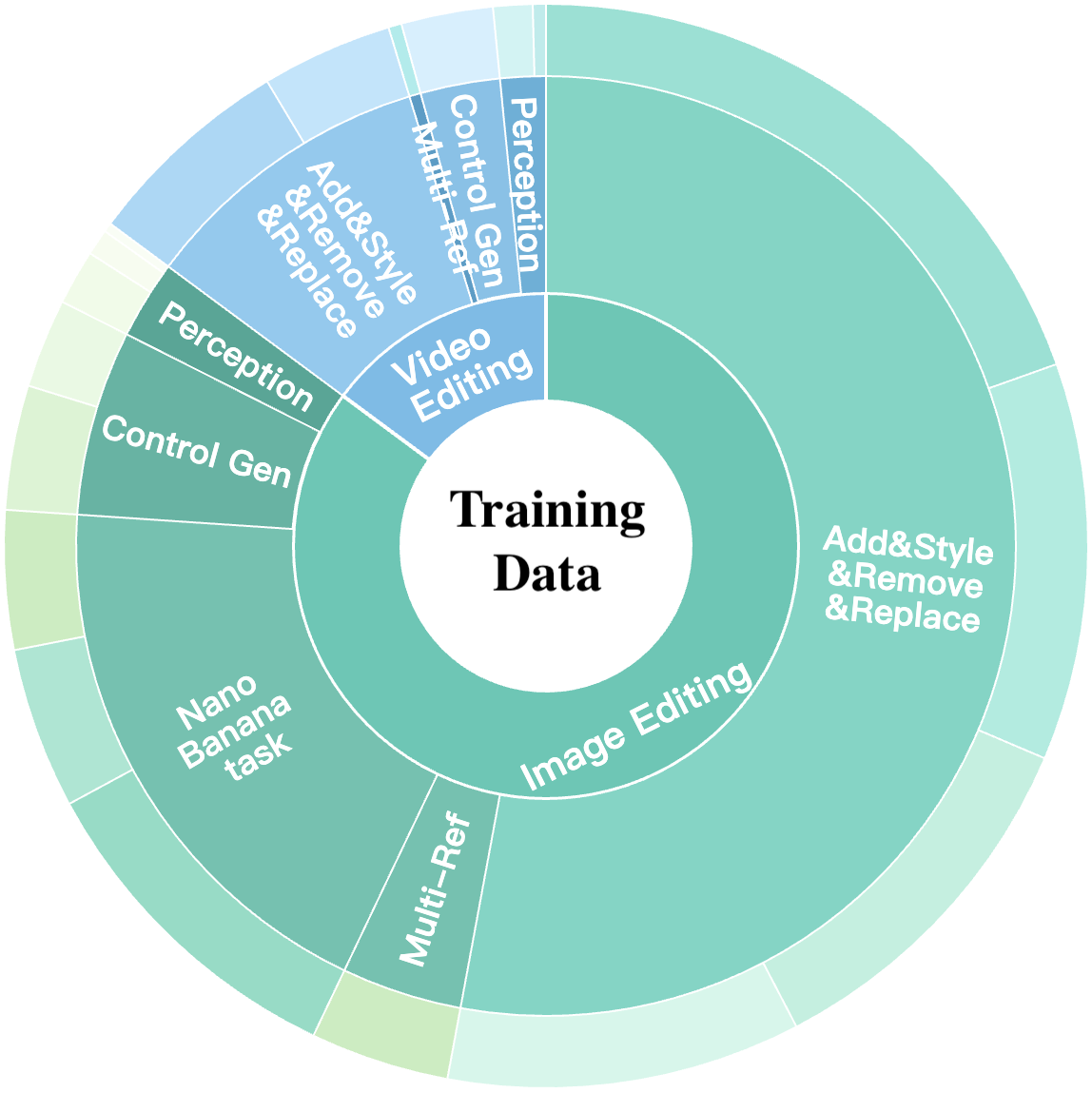} 
    \end{minipage}%
    \hfill
    \begin{minipage}{0.6\textwidth}
        \centering
        \fontsize{6.0pt}{9.0pt}\fontfamily{ptm}\selectfont
        \setlength{\tabcolsep}{1.5pt} 
        \renewcommand{\arraystretch}{1.2} 
        
        \begin{tabular}{ll}
            
            \multicolumn{2}{p{0.98\linewidth}}{\cellcolor{ImgHeaderBG}{\raggedleft\textbf{Image Editing Datasets} \quad \textit{Total: $\sim$2M}}} \\
            
            \ditem{ImgBase!200}{UniWorld-V1~\cite{lin2025uniworld}}{270K}      & \ditem{ImgBase!193}{Pico-banana-400K~\cite{qian2025pico}}{257K} \\
            \ditem{ImgBase!186}{AnyEdit~\cite{jiang2025anyedit}}{280K}       & \ditem{ImgBase!179}{Nano-consistent-150k~\cite{ye2025echo}}{123K} \\
            \ditem{ImgBase!172}{MultiEdit~\cite{li2025multiedit}}{106K}         & \ditem{ImgBase!165}{UnicEdit-10M~\cite{ye2025unicedit}}{500K} \\
            \ditem{ImgBase!158}{Unipic3~\cite{wei2026skywork}}{168K}            & \ditem{ImgBase!151}{ByteMorph~\cite{chang2025bytemorph}}{20K} \\
            \ditem{ImgBase!144}{X2edit~\cite{ma2025x2edit}}{195K}     & \ditem{ImgBase!137}{GPT-Image-Edit-1.5M~\cite{wang2025gpt}}{303K}  \\
            \ditem{ImgBase!130}{RealEdit~\cite{sushko2025realedit}}{42K}           & \ditem{ImgBase!125}{Self-built}{7K} \\

            \multicolumn{2}{p{0.98\linewidth}}{\cellcolor{VidHeaderBG}{\raggedleft\textbf{Video Editing Datasets} \quad \textit{Total: $\sim$380K}}} \\
            
            \ditem{VidBase!200}{Señorita-2M~\cite{zi2025se}}{100K}      & \ditem{VidBase!185}{ReCo-data~\cite{reco}}{160K} \\
            \ditem{VidBase!170}{OpenVE-3M~\cite{he2025openve}}{70K}          & \ditem{VidBase!155}{VideoCoF~\cite{yang2025unified}}{10K} \\
            \ditem{VidBase!140}{Opens2v-nexus~\cite{yuan2025opens2v}}{10K}         & \ditem{VidBase!125}{Self-built}{30K} \\
            
        \end{tabular}
    \end{minipage}
    
    \vspace{-0.5em}
    \caption{Training data statistics and distribution. \textbf{Left:} Distribution of task categories. \textbf{Right:} Statistics of collected datasets. The sampling number for each dataset is displayed.}
    \label{fig:data_statistics}
\end{figure*}

To transfer editing capabilities from the image domain to videos, we curate a massive, cross-modal training dataset. 
We aggregate approximately 2M image editing samples to provide rich spatial and creative priors, coupled with 380K video editing samples to establish temporal consistency.
As illustrated in Fig.~\ref{fig:data_statistics}, while the majority of our data is sourced from public resources, we utilize automated pipelines to synthesize 37K editing pairs for complex tasks, filling capability gaps in existing datasets.

\subsection{Open-source Editing Dataset}
Instead of simply combining datasets, we construct a comprehensive training dataset by aggregating diverse public resources, categorized by input modalities and task types:

\noindent\textbf{1) Single-reference Editing Task.} This category focuses on instruction-based editing with a single reference. For the image domain, we compile 1.85M image-to-image samples from UniWorld-V1~\cite{lin2025uniworld}, AnyEdit~\cite{jiang2025anyedit}, Nano-consistent-150k~\cite{ye2025echo}, MultiEdit~\cite{li2025multiedit}, UnicEdit-10M~\cite{ye2025unicedit}, ByteMorph~\cite{chang2025bytemorph}, GPT-Image-Edit-1.5M~\cite{wang2025gpt}, X2edit~\cite{ma2025x2edit}, RealEdit~\cite{sushko2025realedit}, and Pico-banana-400K~\cite{qian2025pico}. For the video domain, we collect 350K video-to-video samples from Señorita-2M~\cite{zi2025se}, ReCo-data~\cite{reco}, OpenVE-3M~\cite{he2025openve}, and VideoCoF~\cite{yang2025unified}.

\noindent\textbf{2) Multi-reference Editing Task.} To equip the model with the ability to handle multiple reference inputs, we incorporate object insertion tasks from UniWorld-V1~\cite{lin2025uniworld} and AnyEdit~\cite{jiang2025anyedit}. Furthermore, we augment the dataset with 68K multi-reference image generation samples from Unipic3~\cite{wei2026skywork} and 10K multi-reference video generation instances from Opens2v-nexus~\cite{yuan2025opens2v}.

\begin{figure*}[!tbp]
  \centering
   \includegraphics[width=1.0\textwidth]{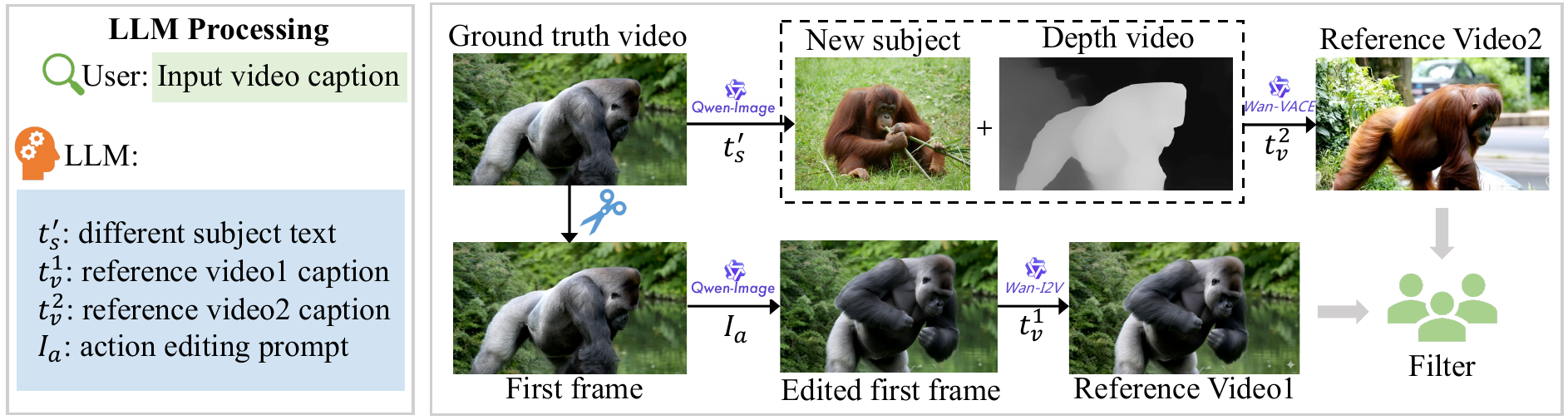} 
  \caption{Overview of the multi-reference action editing data pipeline. Instructions are extracted by an LLM. The action in Reference Video1 is edited to match that of Reference Video2.}
  \label{fig:datapipe}
\end{figure*}

\subsection{Self-built Editing Dataset}
A core bottleneck in current video editing is the lack of data for creative and complex conditional tasks. To bridge this gap, we design targeted, automated synthesis pipelines to generate high-fidelity data tailored for these unexplored scenarios:
1) \textbf{Creative tasks from NBP}: Leveraging the task and instruction list from~\cite{AwesomeNanoBanana2024}, we utilize NBP to synthesize 7,000 creative image editing instances, such as Anime-to-Real Coser transformation and Become-a-VTuber personalization.
We also use Wan-I2V to produce 10K image-to-video data.
2) \textbf{Multi-reference video tasks}: Current multi-condition video models are constrained to video+image or multi-images, leaving the challenging dual-video conditioning scenario (\textit{Reference Video1}, \textit{Reference Video2}, \textit{Target Video}) unexplored. 
To bridge this gap,
taking the multi-reference action editing in Fig.~\ref{fig:datapipe} as an example,
we build automated data pipelines for 5 multi-reference video tasks with 20K samples: 

\textbf{Subject insertion} aims to composite a dynamic object from \textit{Reference Video1} into the scene from \textit{Reference Video2}. 
We start with a real-world video as the ground-truth. 
Then, SAM3~\cite{carion2025sam} is employed to segment the foreground.
The object with a white background is \textit{Reference Video1}. ROSE~\cite{miao2025rose} is applied to remove the subject, yielding the background-only \textit{Reference Video2}.

\textbf{Action editing} transfers the motion of the subject in \textit{Reference Video2} to the subject in \textit{Reference Video1}, while preserving background. 
For \textit{Reference Video2}, we leverage depth sequences within Wan-VACE~\cite{jiang2025vace} to ensure motion consistency. 
For \textit{Reference Video1}, we edit the pose of the first frame with Qwen-Image-Edit~\cite{wu2025qwen}, then employ Wan-I2V~\cite{wan2025wan} to synthesize video.

\textbf{Subject editing} replaces the subject in \textit{Reference Video1} with the subject from \textit{Reference Video2}. 
A real video is considered as \textit{Reference Video1}. 
We use Wan-VACE with subject-controlled inpainting to synthesize the ground truth. 
To construct \textit{Reference Video2}, we edit the background of the initial frame and leverage Wan-I2V for video generation.

\textbf{Camera editing} adapts the camera trajectory of \textit{Reference Video1} to match \textit{Reference Video2}. 
To achieve this, we employ ReCamMaster~\cite{bai2025recammaster} to create videos with different camera motions but same content.

\textbf{Stylization} transfers the style of \textit{Reference Video2} to \textit{Reference Video1}.
We stylize the first frame and leverage depth-conditioned generation in Wan-VACE to produce different stylized videos.

\subsection{Data Post-processing}
We implement post-processing for the aggregated dataset, focusing on three key aspects:
1) Instruction Standardization: We normalize editing
instructions across different datasets into a unified format to ensure consistency.
2) Instruction Reformulation: For tasks such as reference image generation or controllable video generation, where only target video captions are available, we rewrite them into explicit editing instructions.
3) Synthetic Data Filtering: Given the challenges MLLM face in simultaneously comprehending triple-video inputs, we employ manual verification to filter our synthesized multi-reference video editing data.

\section{Methodology}

\begin{figure*}[!tbp]
  \centering
   \includegraphics[width=1.0\textwidth]{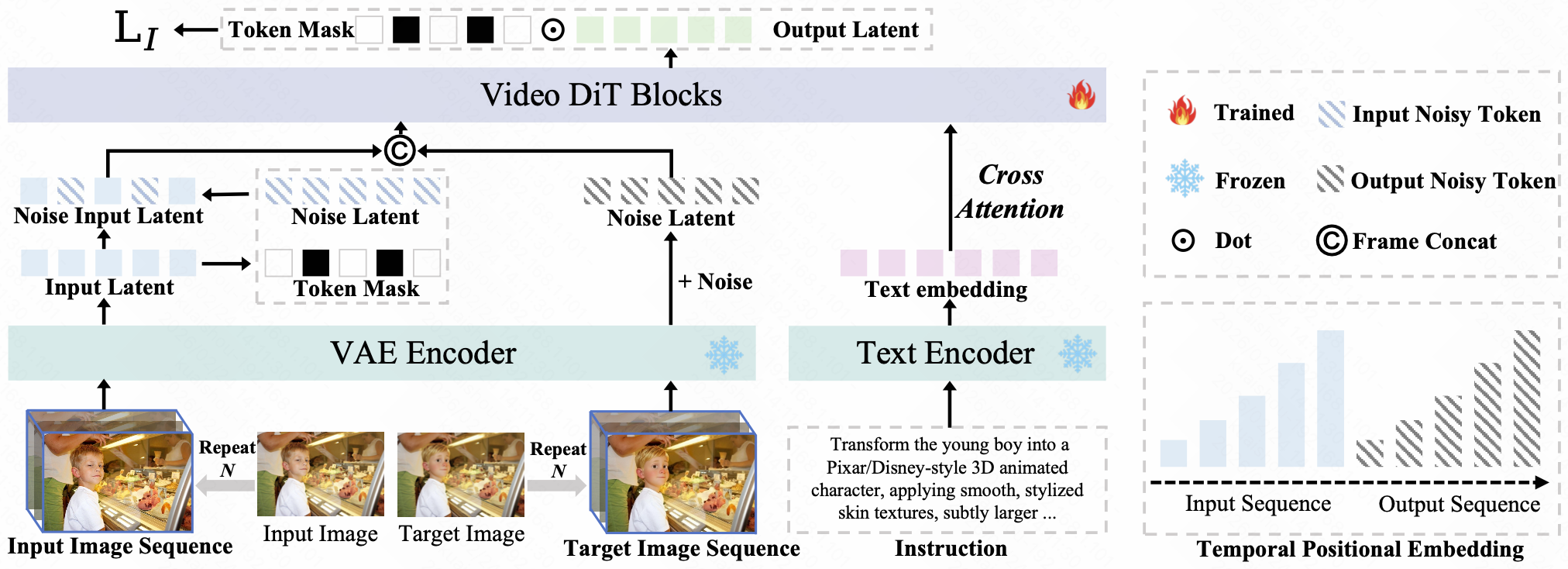} 
  \caption{Overview of the LIVE pipeline. For image data, the input and target images are replicated $N$ times to construct pseudo-video sequences, whereas for video data, the latents are extracted directly. The input latents and noise latents are then concatenated along the temporal dimension.}
  \label{fig:pipeline}
\end{figure*}

\subsection{Overview}
In the context of single-reference instruction-based video editing, given a reference video $V_{ref} \in \mathbb{R}^{f \times c \times h \times w}$ and a textual instruction $C$, the objective is to generate a target video $V_{tar} \in \mathbb{R}^{f_t \times c \times h_t \times w_t}$ that follows the instruction while preserving the essential content of $V_{ref}$.
As shown in Fig.~\ref{fig:pipeline}, we encode the $V_{ref}$ into latent representations $z_{ref}$ using the 3D VAE encoder in~\cite{wan2025wan}, while $C$ is tokenized into sequences $z_{c}$ with the T5 tokenizer~\cite{2020t5}.
During training, we perturb the latent representation of 
 $V_{tar}$ with random noise to obtain the noisy latent 
$z_{tar}$.
Video DiT from~\cite{wan2025wan} is employed as the backbone of our method.
To facilitate robust temporal scalability and accommodate multi-reference scenarios, we concatenate the reference video tokens and noisy target tokens along the temporal dimension to form a unified sequence $z = \left[ z^1_{\text{ref}}, \dots, z^n_{\text{ref}}, z_{\text{tar}} \right]$, where $z^n_{\text{ref}}$ denotes the token sequence of the $n$-th reference video.
$z_{c}$ is integrated into the model via cross-attention mechanisms.

\textbf{Preliminaries.} To ensure training stability, we adopt Flow Matching as our foundational theoretical framework. This approach leverages ordinary differential equations (ODEs) while maintaining equivalence to maximum likelihood objectives.
Given a video latent $x_{1}$, a Gaussian noise sample $x_{0} \sim \mathcal{N}(0, I)$, and a timestep $t \in[0,1]$ sampled from a logit-normal distribution, the intermediate latent is defined as $x_t=t x_1 + (1 - t) x_0$.
The ground truth velocity field $v_{t}$ is given by:
\begin{equation}\label{v_t}
v_{t}=\frac{d x_{t}}{d t}=x_{1}-x_{0}.
\end{equation}
The training objective minimizes the mean squared error between the model prediction $f$ (parameterized by $\theta$) and $v_{t}$:
\begin{equation}
\mathcal{L}_{v} = \mathbb{E}_{x_0, x_1, I, t}||f(x_t, I, t; \theta)-v_t||^2,
\end{equation}

\subsection{Video-Image Alignment}
\textbf{Image Repeat.} 
Training Video DiT directly on a mixture of large-scale image and video datasets introduces significant inconsistency along the temporal dimension. 
This discrepancy causes the model to focus solely on single-frame spatial transformations while neglecting temporal dynamics. 
To mitigate this issue, we employ an image repetition strategy to align the data. 
Specifically, given a source image $I_r$, a target image $I_t$, and an editing instruction $C_I$, we replicate both $I_r$ and $I_t$ $N$ times along the temporal axis, where $ N$ is set to 49.
This process constructs pseudo-video sequences $V_r \in \mathbb{R}^{N \times c \times h_r \times w_r}$ and $V_t \in \mathbb{R}^{N \times c \times h_t \times w_t}$, ensuring aligned temporal dimensionality across all training inputs.

\textbf{Frame-wise Token Noise.} While image repetition unifies data formats, naive repetition introduces a static bias, preventing the model from learning dynamic transitions. 
This is particularly detrimental when video data of a certain task is scarce, as the model tends to overfit to the identity mapping between frames.
Inspired by~\cite{yang2025unified,wu2025chronoedit}, we propose a Frame-wise Token Noise strategy. 
Given the pseudo-video latent representation $z_I$ from the 3D VAE encoder, we generate a 1D temporal binary mask $M \in \{0, 1\}^l$ based on a masking ratio $R$, $l$ is the temporal length of $z_I$.
Specifically, for each index $i \in \{1, \dots, l\}$, we replace its corresponding spatial token sequence in $z_I$ with Gaussian noise if $M_i = 1$.
Crucially, during the calculation of the flow matching loss, we exclude these noisy tokens from optimization. The modified objective is formulated as:
\begin{equation}
\mathcal{L}_{I} = \mathbb{E}_{x_0, x_1, I, t} \left[ (1 - M) \odot ||f(x_t, I, t; \theta)-v_t||^2 \right].
\end{equation}
By setting the loss of masked tokens to zero, we prevent the model from being penalized for deviating from the static ground truth. 
This encourages the model to implicitly reason about intermediate temporal content and bridge the gap between static frames with potential dynamics.

\textbf{Temporal RoPE.} 
Let $L_i$ and $L_t$ denote the temporal lengths of the input video and the noisy target latent, respectively. 
Video DiT typically applies sequential temporal indices, i.e., $[0, \dots, L_i + L_t - 1]$ across the concatenated token sequence. 
To facilitate video length extrapolation, we adopt a reset configuration strategy following~\cite{yang2025unified}. 
Specifically, we decouple the temporal indices of the reference and target videos, assigning indices $[0, \dots, L_i-1]$ and $[0, \dots, L_t-1]$.
It is worth noting that we explicitly assign valid temporal progression indices to the noisy tokens. 
This preserves the temporal structure, allowing the model to effectively reason about the underlying temporal dynamics during generation.

\subsection{Two-stage Training Strategy}
To effectively transfer the editing capabilities learned from large-scale static image datasets to the video domain, we employ a two-stage curriculum learning strategy.
We initiate training with a mixed dataset characterized by a high image-to-video ratio, comprising 2M images and 350K videos. 
This phase focuses primarily on single-reference instruction editing tasks, allowing the model to establish a strong foundation in content manipulation.
In the subsequent stage, we rebalance the data distribution to facilitate the learning of temporal dynamics.
We adjust the sampling ratio to increase the relative proportion of video data, resulting in a dataset of 116K videos and 260K images. 
Additionally, we incorporate more complex multi-reference editing tasks during this phase to enhance task diversity and model robustness.
\section{Experiment}
\begin{table}[t]
\centering
\caption{Quantitative comparison on LIVE-Bench. Basic, Nano, Control, Percept, Multi-Ref are add \& style \& remove \& replace, creative tasks from Nano Banana, controllable generation, perception, multi-reference editing tasks.}
\resizebox{0.98\textwidth}{!}{
\begin{tabular}{ll|cc|c|cc|cc}
\toprule
\toprule
\multirow{3}{*}{\textbf{Method}} & \multirow{3}{*}{\textbf{Type}} & \multicolumn{2}{c|}{\textbf{VLM evaluation}} & \multicolumn{1}{c|}{\textbf{Video Quality}} & \multicolumn{2}{c|}{\textbf{Text Alignment}} & \multicolumn{2}{c}{\textbf{Temporal Consistency}} \\ 
\cmidrule{3-4}\cmidrule{5-5}\cmidrule{6-7}\cmidrule{8-9}
& & \textbf{Edit. Qual. $\uparrow$} & \textbf{Task SR $\uparrow$} & \textbf{Pick Score $\uparrow$} & \textbf{Frame $\uparrow$} & \textbf{Video $\uparrow$} & \textbf{CLIP $\uparrow$} & \textbf{DINO $\uparrow$} \\
\midrule

\multirow{6}{*}{\textbf{Lucy Edit}~\cite{decart2025lucyedit}} 
& Basic & 7.16 & 72.88 & 19.60 & 29.44 & 22.37 & 98.84 & \textbf{98.77} \\
& Nano & 5.59 & 39.56 & 19.35 & 27.62 & 18.26 & 98.40 & 98.31 \\
& Control & 7.65 & 77.27 & 19.84 & 29.17 & 22.76 & 97.54 & 97.90 \\
& Percept & 5.42 & 31.25 & \textbf{20.02} & \textbf{24.99} & \textbf{20.44} & 98.62 & 98.52 \\
& Multi-Ref & - & - & - & - & - & - & -  \\
& Overall & 5.98 & 47.26 & 19.45 & 27.89 & 19.38 & 98.47 & 98.40 \\
\midrule
\multirow{6}{*}{\textbf{Ditto}~\cite{bai2025scaling}} 
& Basic & 7.03 & 69.83 & 20.24 & 30.04 & 21.68 & \textbf{98.94} & 98.91 \\
& Nano & 5.93 & 52.78 & 19.66 & 26.65 & 16.99 & 98.81 & 98.66 \\
& Control & 8.81 & 100.00 & 20.88 & 29.38 & 25.09 & 98.57 & 98.73 \\
& Percept & 4.66 & 35.14 & 19.79 & 24.58 & 19.45 & 98.65 & 98.79 \\
& Multi-Ref & - & - & - & - & - & - & - \\
& Overall & 6.17 & 43.44 & 19.83 & 27.29 & 18.36 & 98.81 & 98.72 \\
\midrule
\multirow{6}{*}{\textbf{ReCo}~\cite{reco}} 
& Basic & 6.82 & 66.12 & 19.71 & 28.45 & 20.75 & 98.39 & 98.83 \\
& Nano & 5.33 & 36.23 & 19.27 & 26.67 & 17.55 & 97.95 & 98.12 \\
& Control & 7.41 & 82.61 & 20.27 & 29.27 & 23.31 & 97.78 & 98.62 \\
& Percept & 3.77 & 0.00 & 18.66 & 22.67 & 15.58 & 98.60 & \textbf{99.01} \\
& Multi-Ref & - & - & - & - & - & - & - \\
& Overall & 5.70 & 43.84 & 19.39 & 27.09 & 18.43 & 98.05 & 98.31 \\
\midrule
\multirow{6}{*}{\textbf{Ours}} 
& Basic & \textbf{8.56} & \textbf{93.39} & \textbf{20.38} & \textbf{30.87} & \textbf{24.25} & 98.90 & 98.74 \\
& Nano & \textbf{7.96} & \textbf{85.51} & \textbf{20.28} & \textbf{29.72} & \textbf{21.17} & \textbf{99.01} & \textbf{98.77} \\
& Control & \textbf{8.91} & \textbf{100.00} & \textbf{21.11} & \textbf{29.96} & \textbf{24.72} & \textbf{98.96} & \textbf{98.92} \\
& Percept & \textbf{5.99} & \textbf{44.00} & 19.76 & 24.09 & 17.72 & \textbf{98.74} & 98.47 \\
& Multi-Ref & \textbf{5.82} & \textbf{49.02} & \textbf{19.61} & \textbf{22.97} & \textbf{17.25} & \textbf{99.05} & \textbf{99.07} 
\\
& Overall & \textbf{7.79} & \textbf{81.49} & \textbf{20.23} & \textbf{28.99} & \textbf{21.29} & \textbf{98.97} & \textbf{98.77} \\
\bottomrule
\end{tabular}
}
\label{tab:our_bench}
\end{table}

\begin{table}[t]
\centering
\vspace{-15pt}
\caption{Quantitative comparison on EditversBench. The results of Lucy Edit and EditVerse are from~\cite{ju2025editverse}.}
\resizebox{0.98\textwidth}{!}{
\begin{tabular}{l|c|c|cc|cc}
\toprule
\toprule
\multirow{2}{*}{\textbf{Method}} & \multicolumn{1}{c|}{\textbf{VLM evaluation}} & \multicolumn{1}{c|}{\textbf{Video Quality}} & \multicolumn{2}{c|}{\textbf{Text Alignment}}  &  \multicolumn{2}{c}{\textbf{Temporal Consistency}} \\ \cmidrule{2-7} 
 & \textbf{Editing Quality $\uparrow$}   & \textbf{Pick Score $\uparrow$}    & \textbf{Frame $\uparrow$}            & \textbf{Video $\uparrow$}      & \textbf{CLIP $\uparrow$}                & \textbf{DINO $\uparrow$}                                         \\ \midrule
\midrule
 \textbf{Lucy Edit}~\cite{decart2025lucyedit}    & 5.89     & 19.67  & 26.00 & 23.11  &   98.49 & 98.38    \\
  \textbf{Ditto}~\cite{bai2025scaling}    & 5.99     & 19.30  & 23.83 & 18.93  &   98.55 & \textbf{98.99}    \\
  \textbf{ReCo}~\cite{reco}    & 6.60     & 19.01  & 24.71 & 20.69  &   98.08 & 98.28    \\
\textbf{EditVerse}~\cite{ju2025editverse}  &7.65        &   \textbf{20.07}   &26.73             &    23.93    &98.56              &     98.42    \\
  \textbf{Ours}  &\textbf{8.01} &20.01 &\textbf{27.11} &\textbf{24.07}  &\textbf{98.73} &98.42 \\ 
\midrule
\end{tabular}
}
\label{tab:editverse_bench}
\end{table}

\subsection{Implementation Details}
We utilize Wan2.2-5B-TI2V~\cite{wan2025wan} as the initial weight of DiT. 
To accommodate dynamic resolutions, we resize all videos and images such that the shorter side is 640 pixels, while maintaining the aspect ratio and ensuring that dimensions are multiples of 32. 
During training, Video DiT is fully fine-tuned, while all other components remain frozen.
We employ a learning rate of $1 \times 10^{-4}$ and a weight decay of 0.01. 
The token masking ratio is set to 25\%. Experiments are conducted on 64 GPUs using the PyTorch framework with a global batch size of 128. The training process consists of two stages: we train for 33,000 steps in the first stage and 6,000 steps in the second stage.

\subsection{Evaluation Settings}
\noindent\textbf{Testing benchmark.} 
Early video evaluation benchmarks mainly focus on assessing training-free methods, like V2VBench~\cite{sun2024diffusion} and TGVE~\cite{singer2024video}. 
More recently, several studies have proposed benchmarks designed for instruction-based video editing, such as Ivebench~\cite{chen2025ivebench} and OpenVE Bench~\cite{he2025openve}.
However, their task diversity remains limited.
While EditVerseBench~\cite{ju2025editverse} introduces a dataset comprising over 20 tasks, these are predominantly concentrated on fundamental editing operations.
To address these limitations, we propose \textbf{LIVE-Bench}, which is designed for evaluating a more diverse range of editing tasks and the ability to transfer image editing knowledge. 
LIVE-Bench is structured into several key components:
1) Basic Editing Tasks: These include global editing, local editing, perception editing, and controllable generation.
2) Image-to-Video Transfer Tasks: These encompass tasks absent in existing video editing datasets but are well-established in image editing datasets.
3) Complex Context-Aware Editing: This category features intricate context-dependent editing tasks, as demonstrated by NBP.
This benchmark comprises 600 samples covering over 60 distinct subtasks. 
All source videos are collected from Pixabay, while the editing instructions and video caption prompts are generated using Gemini 3. 
Furthermore, to facilitate comparison with existing works, we also conduct evaluations on the public \textbf{EditVerseBench}~\cite{ju2025editverse}.

\textbf{Comparison methods.}
Our evaluation includes a comparison with leading open-source end-to-end video editing frameworks: Lucy Edit~\cite{decart2025lucyedit}, Ditto~\cite{bai2025scaling}, and ReCo~\cite{reco}. Furthermore, we compare with EditVerse on EditVerseBench, which is trained jointly on both video and image datasets.

\textbf{Evaluation metrics.} Following  ~\cite{ju2025editverse}, we evaluate performance across four dimensions: VLM evaluation, Video Quality, Text Alignment, and Temporal Consistency.
Additionally, to assess the diversity of supported tasks, we report the Task Success Rate (Task SR). Given a maximum Editing Quality score in VLM evaluation of 9, Task SR is defined as the proportion of tasks that achieve an average score greater than 6.

\subsection{Comparison with the State-of-the-art Methods}
To verify the effectiveness of the proposed method, we conduct quantitative and qualitative comparisons on LIVE-Bench and EditVerseBench.

\begin{table}[t]
\centering
\caption{Ablation study on LIVE-Bench to prob components proposed in LIVE.}
\resizebox{0.98\textwidth}{!}{
\begin{tabular}{cccc|cc|c|cc|cc}
\toprule
\toprule
\multicolumn{4}{c|}{\textbf{Modules}} & \multicolumn{2}{c|}{\textbf{VLM evaluation}} & \multicolumn{1}{c|}{\textbf{Video Quality}} & \multicolumn{2}{c|}{\textbf{Text Alignment}}  &  \multicolumn{2}{c}{\textbf{Temporal Consistency}} \\ 
\cmidrule{1-4} \cmidrule{5-6} \cmidrule{7-7} \cmidrule{8-9} \cmidrule{10-11}
\textbf{Image} & \textbf{Repeat} & \textbf{Noise} & \textbf{Stage2} & \textbf{Edit. Quality $\uparrow$} & \textbf{Task SR $\uparrow$}  & \textbf{Pick Score $\uparrow$}    & \textbf{Frame $\uparrow$}            & \textbf{Video $\uparrow$}      & \textbf{CLIP $\uparrow$}                & \textbf{DINO $\uparrow$}                                         \\ \midrule
$-$ & $-$ & $-$ & $-$ & 6.65 & 62.67 & 19.67 & 26.85 & 19.02 & 98.62  & 98.37\\ 
$\checkmark$ & $-$ & $-$ & $-$ & 7.01 & 71.01 & 20.01 & 27.55 & 19.98 & 98.40   & 98.11 \\ 
$\checkmark$ & $\checkmark$ & $-$ & $-$ & 7.35 & 80.56 & 20.11 & 28.12 & 20.70 & 98.65  & 98.15 \\ 
$\checkmark$ & $\checkmark$ & $\checkmark$ & $-$ & 7.43 & 81.23 & 20.19 & 28.80 & 21.00 & 98.65  & 98.51 \\  
$\checkmark$ & $\checkmark$ & $\checkmark$ & $\checkmark$ &   \textbf{7.79} & \textbf{81.49} & \textbf{20.23} & \textbf{28.99} & \textbf{21.29} & \textbf{98.97} & \textbf{98.77} \\ 
\bottomrule
\end{tabular}
}
\label{tab:ab}
\end{table}

\begin{figure*}[t]
  \centering
  \vspace{-10pt}
   \includegraphics[width=1.0\textwidth]{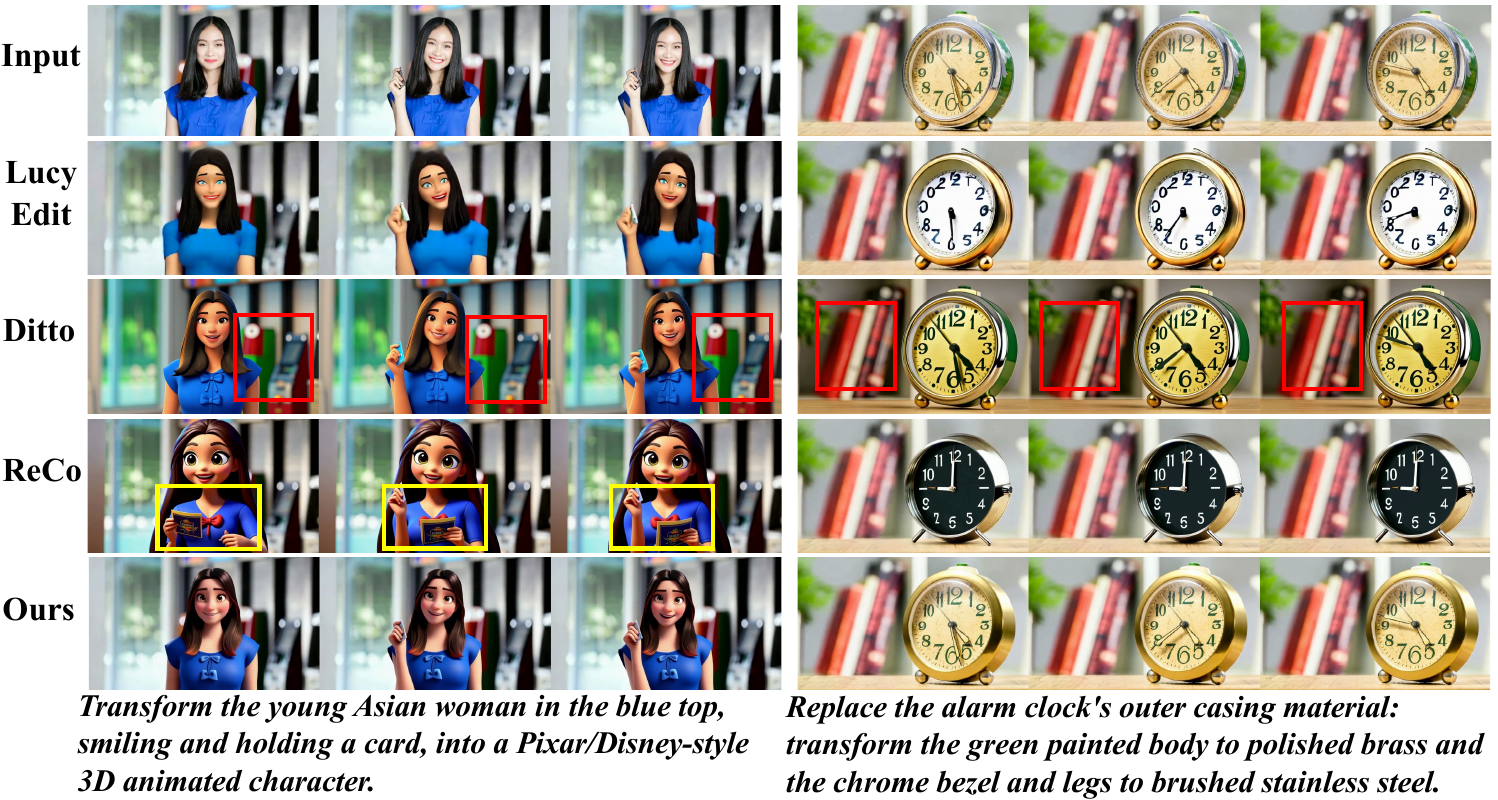} 
  \caption{Qualitative comparison with open-source instruction-based video editing models on LIVE-Bench. LIVE demonstrates superior capabilities in instruction following, background preservation, and temporal consistency.}
  \label{fig:vis_com}
\end{figure*}

\textbf{Quantitative comparison.}
Firstly, the results of LIVE-Bench are shown in Tab.~\ref{tab:our_bench}. 
Overall, despite utilizing only 380K video editing samples, our method achieve state-of-the-art results across all 7 metrics.
Notably, the improvements in Editing Quality and Task SR are most prominent, with gains of 1.62 and 34.23, respectively.
These results validate that video editing knowledge can be effectively transferred to image domains within our training framework.
Moreover, LIVE maintains the best temporal consistency even though it is trained on a large corpus of static image data. 
This is attributed to our frame-wise token noise strategy, which effectively constructs dynamic pseudo-videos enabling the model to learn temporal dependencies.
Furthermore, we also achieve the best performance on EditVerseBench. 
Notably, despite utilizing less video data compared to EditVerse's 4.2M, we obtain superior VLM evaluation score, demonstrating the effectiveness of our approach in temporal preservation.

\textbf{Qualitative comparison.}
The qualitative results are shown in Fig.~\ref{fig:vis_com}. 
The superiority of our proposed method is mainly due to two aspects:
1) Content Preservation. First, benefiting from the extensive image editing priors, our method precisely preserves spatial information. 
For instance, as shown in the left example, while Ditto (indicated by the red box) inadvertently degrades the background, our approach successfully applies stylization to the Asian woman while leaving the background intact.
Second, owing to our data alignment strategy, LIVE demonstrates superior temporal preservation. 
In the left example, while ReCo generates human motion that is inconsistent with the input video (yellow box), our approach faithfully preserves the original action.
In the right example, the motion trajectory of the alarm clock hands is also perfectly maintained in LIVE.
2) Instruction Following. While competing methods like Ditto exhibit reasonable temporal stability, they often fail to strictly adhere to editing instructions. 
In the alarm clock case, Ditto fails to alter the material as requested. 
In contrast, our method successfully executes the material change while simultaneously preserving temporal consistency.

\begin{figure*}[!tbp]
  \centering
   \includegraphics[width=1.0\textwidth]{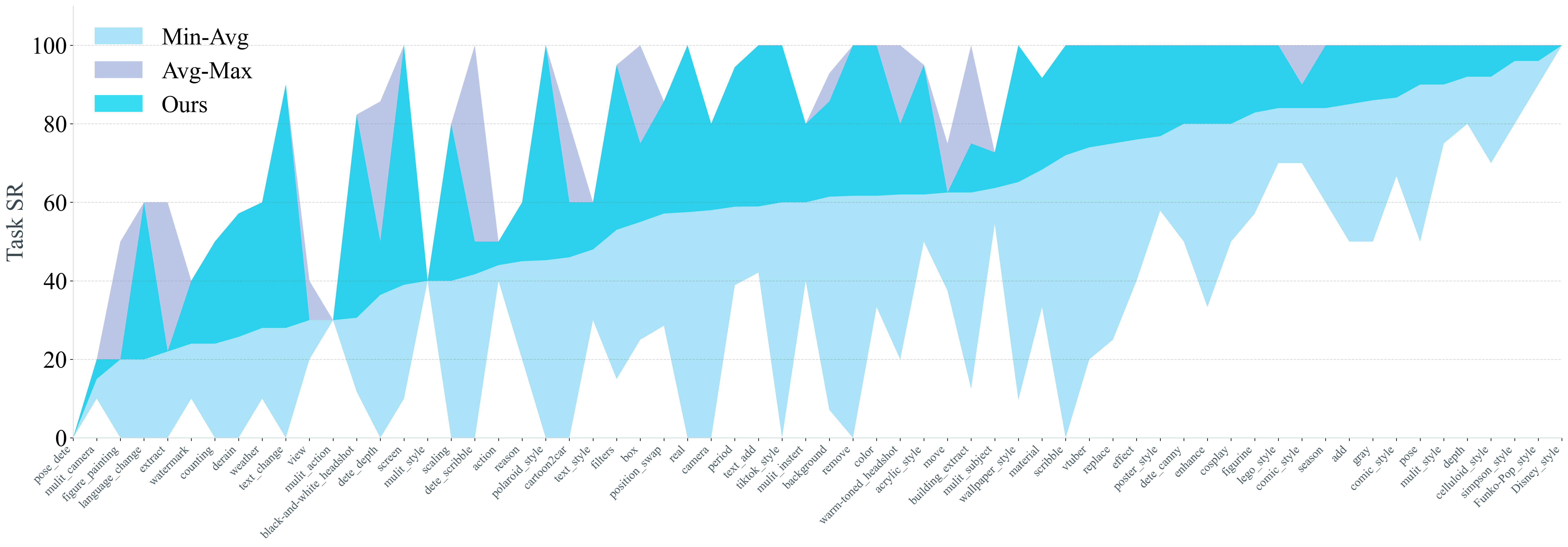} 
  \vspace{-20pt}
  \caption{Per-task performance on LIVE-Bench with Task SR. Tasks are arranged based on the average performance of all comparison methods and baseline trained with only video data. We visualize the minimum, average, and maximum scores for each task, compared with our method. We achieve the highest performance on 52 of 64 tasks.}
  \vspace{-5pt}
  \label{fig:task_res}
\end{figure*}

\textbf{Per-task performance.}
As shown in Fig.~\ref{fig:task_res}, beyond the overall quantitative evaluation, we present a detailed performance breakdown across individual tasks. 
Specifically, we visualize the minimum, average, and maximum scores alongside the results achieved by our proposed method.
Notably, our approach achieves the highest scores for the majority of tasks, \ i.e., 52, demonstrating its robustness in diverse editing scenarios.
Some tasks have low overall performance.
It can be attributed to two core factors. 
First, the sample availability for certain tasks is limited. 
For instance, the text language editing task contains only 1897 samples among our 2M training data.
Data imbalance is prone to inducing catastrophic forgetting during model training, thereby leading to performance degradation. 
Second, the model lacks essential supervision of temporal knowledge. 
Particularly for complex tasks like multi-reference camera movement editing, the required dynamic camera features from reference videos are poorly conveyed by static images, making effective cross-modal knowledge transfer highly challenging.
Addressing these issues to further enhance performance on these challenging tasks remains a promising avenue for future research.

\subsection{Ablation Study}
\textbf{Effectiveness of the proposed components.}
As shown in Tab.~\ref{tab:ab}, we conduct ablation experiments on our proposed module on LIVE-Bench.
1) Although the incorporation of image data improve instruction following, it led to a degradation in temporal consistency.
2) The first-stage training yield significant improvements. 
Specifically, the repeat strategy aligned data dimensions, enhanced training stability, and accelerated the acquisition of editing instructions. 
The token noise strategy further preserved temporal information during mixed-data training, improving video quality and temporal consistency.
3) The second stage, trained at an increased video-to-image ratio, not only enhance complex tasks but also restore temporal consistency.
4) By combining all modules, the model achieves the best overall results, demonstrating the complementarity of our domain alignment and curriculum learning strategies.

\begin{figure*}[!tbp]
  \centering
   \includegraphics[width=1.0\textwidth]{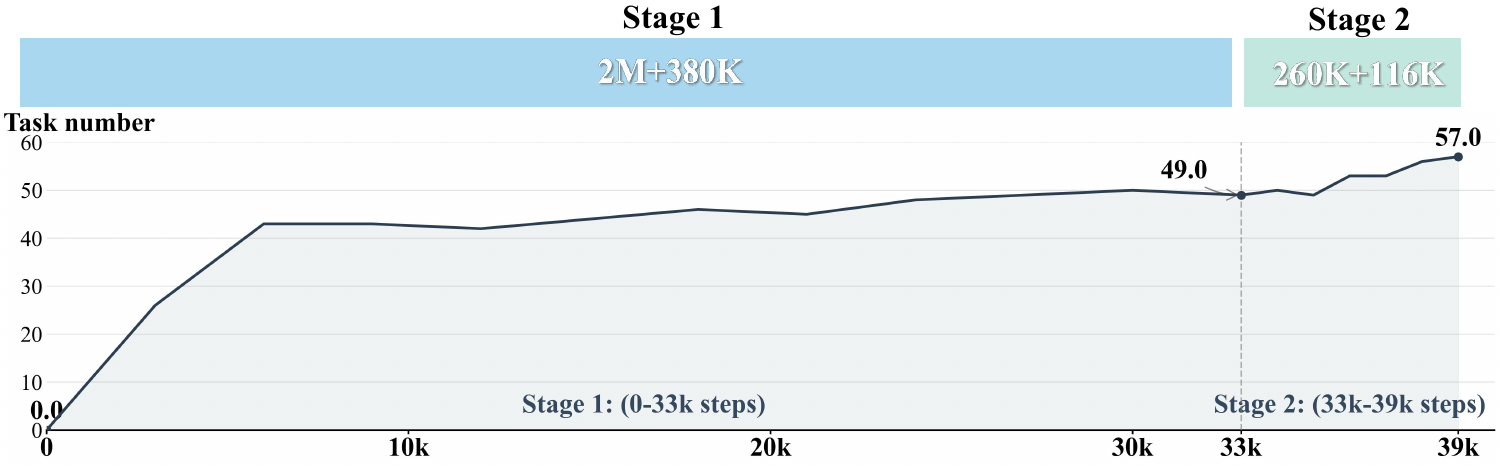} 
  \caption{The dynamic training process of LIVE. The number of tasks with an average Editing Quality score exceeding 6 in different training steps is counted. In the first stage, 2M image data and 380K video data are trained. In the second stage, the proportion of images and videos is reduced, \ i.e., 260K video and 116K image. Moreover, more challenging tasks are included.}
  \label{fig:steps}
\end{figure*}

\textbf{Two-stage training.} 
To validate the effectiveness of the two-stage training strategy, we visualize supported tasks number of LIVE across different training steps in Fig.~\ref{fig:steps}. 
Specifically, a task is considered supported if the average Editing Quality score across all its samples exceeds 6.
In the first training stage, the model acquires knowledge of numerous and diverse editing tasks from extensive image data, enabling a rapid increase in the number of supported tasks at an early training phase.
Nevertheless, during subsequent long-term training, the growth in the number of supported tasks slows down considerably, and even performance fluctuations arise.
On the one hand, the model inevitably suffers from static bias, leading to inferior overall quality of the edited videos.
On the other hand, several challenging tasks suffer from limited sample availability, making it difficult for the model to maintain consistent performance over prolonged training iterations.
Accordingly, in the second training stage, we increase the proportion of video samples, encouraging the model to focus on learning temporal dependencies.
Concurrently, more challenging editing tasks (e.g., multi-reference video editing) are introduced, and a curriculum-based knowledge injection strategy is adopted to progressively impart complex knowledge, leading to more stable training.
Empirically, the number of editable tasks supported by the model is boosted from 49 to 57 using merely 260K image samples and 116K video samples in this stage.

\begin{figure*}[!tbp]
  \centering
   \includegraphics[width=1.0\textwidth]{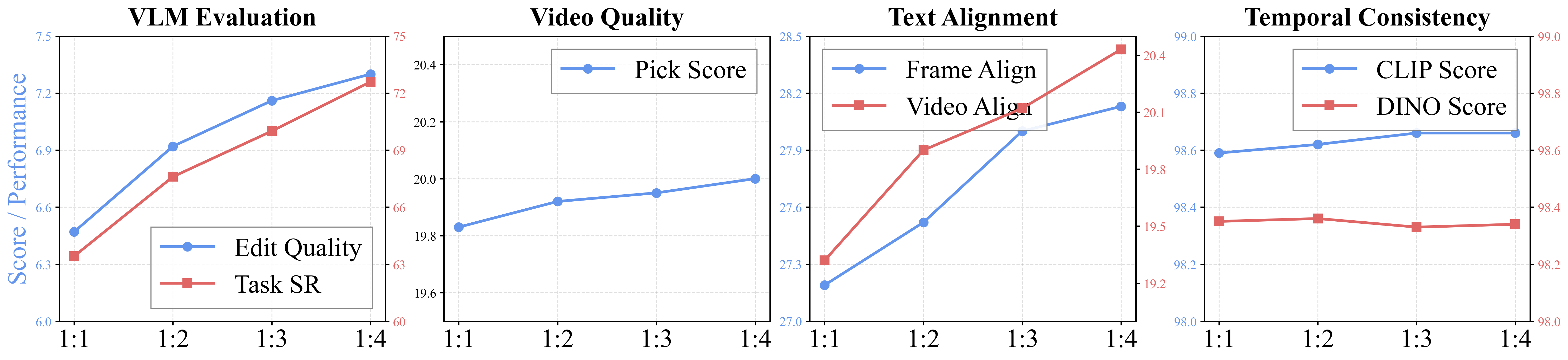} 
  \caption{The ablation study on the video-image mixing ratio. We conduct experiments under the settings of 1:1, 1:2, 1:3, and 1:4.}
  \vspace{-10pt}
  \label{fig:scaling}
\end{figure*}
\textbf{Scaling law.}
To validate the scaling law of our training framework, we conduct an empirical analysis scaling the volume of integrated image data. 
Specifically, we establish a baseline using a sampled video-editing subset of 22K instances across tasks, and progressively augment it with 22K, 44K, 66K, and 88K image editing samples. 
The results, illustrated in Fig.~\ref{fig:scaling}, reveal the following key observations:
1) The model exhibits continuous enhancements in both editing quality and text alignment as the scale of image data increases. 
This upward trend substantiates that the framework enables image-editing knowledge to be efficiently transferred to the video domain.
2) In contrast to the rapid gains in editing quality, improvements in video quality are relatively marginal. 
We attribute this to the scarcity of video data in certain editing tasks. 
Lacking explicit temporal supervision, it remains highly challenging for the model to synthesize complex motion changes in a zero-shot manner.
We hypothesize that integrating a more powerful backbone with richer pre-trained priors could be a promising avenue to alleviate this limitation in future work.
3) Finally, temporal consistency remains robust throughout the scaling. The introduction of massive image data does not induce any temporal collapse, further demonstrating that our method ensures training stability and effectively preserves crucial dynamics.

\section{Conclusion}
In this paper, we propose LIVE for training instruction-based video editing with image editing knowledge.
To align the dimensions of image and video editing data, we first perform temporal replication. 
To mitigate the static bias inherent in naive repetition, we introduce a frame-wise token noise strategy, which treats the latents of specific frames as generative priors to preserve temporal information.
Furthermore, we employ a two-stage curriculum learning, progressively incorporating more challenging tasks to enhance task diversity while gradually reducing the image-to-video sample ratio to restore temporal consistency. 
We also establish LIVE-Bench, a comprehensive benchmark consisting of over 60 sub-tasks that encompass rare scenarios absent in video datasets but prevalent in image domains.
Extensive comparative and ablation studies demonstrate the effectiveness of our proposed method.

\bibliographystyle{splncs04}
\bibliography{main}
\end{document}